# 6$^{th}$ Grid-Net: Unified Remote Sensing Image Dehazing Based on Color Restoration and Edge-Preserving

Runci Bai , Kui Jiang , Xiang Chen , Chen Wu , Dianjie Lu , Guijuan Zhang , Zhuoran Zheng

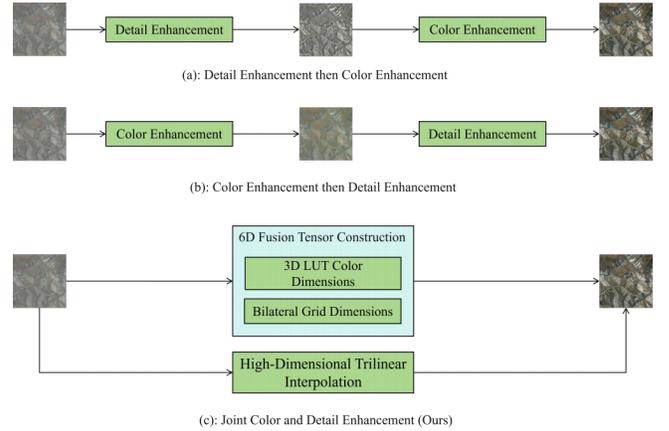

Fig. 1. Comparison of remote sensing image quality restoration paradigms: (a) Detail enhancement followed by color rendition. (b) Color rendition followed by detail enhancement. (c) The proposed 6$^{th}$ Grid-Net collaboratively performs color rendition and detail enhancement via a six-dimensional fusion tensor (constructed from 3D LUT color dimensions and bilateral grid spatial-luminance dimensions) and manifold-adaptive high-dimensional sampling. Our method optimizes color stylization and edge preservation simultaneously, circumventing the mutual interference and artifact accumulation inherent in sequential pipelines.

*Abstract*—Remote sensing images are frequently degraded by adverse weather conditions, particularly clouds and haze, which severely impair downstream applications. Existing restoration methods typically rely on computationally heavy architectures or sequential pipelines (*e.g.*, detail enhancement followed by color rendition) that suffer from mutual interference and artifact accumulation. Furthermore, recent unified grid-based approaches utilize fixed, isotropic interpolation kernels, neglecting the intrinsic low-dimensional manifold of natural images and inevitably causing edge blur. To address these limitations, we propose 6$^{th}$ Grid-Net, a highly efficient and unified remote sensing image restoration framework tailored for resource-constrained edge devices. Specifically, we construct a novel six-dimensional fusion tensor that seamlessly integrates the color rendition capabilities of 3D LUTs with the spatial-luminance detail preservation of bilateral grids. To overcome the drawbacks of standard trilinear interpolation, we introduce a manifold-adaptive high-dimensional sampling mechanism. This mechanism dynamically adjusts the interpolation kernel based on local edge orientation, texture strength, and color similarity, enabling simultaneous global color stylization and local edge refinement in a single forward pass. Additionally, an edge-aware grid smoothing constraint and dynamic quantization are incorporated to suppress ghosting artifacts and significantly compress the model size. Extensive experiments on multiple benchmark datasets demonstrate that 6$^{th}$ Grid-Net achieves state-of-the-art restoration quality across various degradation scenarios. Moreover, our lightweight model exhibits strong generalization in natural scene colorization and supports real-time processing of 1080P high-resolution images on consumer-grade mobile devices.

## I. INTRODUCTION

REMOTE sensing images are crucial for surface information acquisition, underpinning applications such as agricultural assessment, urban planning, and climate change research. However, these images are frequently degraded by adverse weather conditions during acquisition, with clouds and haze being the predominant degradation factors. Clouds typically obscure surface information and introduce complex scattering effects, whereas haze reduces image contrast, causes color shifts, and blurs high-frequency details. The distinct physical characteristics of these two degradations pose a significant challenge for developing a unified restoration algorithm. Furthermore, the inherently high resolution of remote sensing images (*e.g.*, $\geq 512 \times 512$) imposes strict real-time processing requirements on such algorithms.

Traditional remote sensing image restoration methods are typically tailored to a single degradation type, struggling to simultaneously and effectively handle both clouds and haze. Recently, "all-in-one" joint restoration methods have emerged, attempting to address multiple degradations within a unified model. By stacking deep convolutional layers, incorporating self-attention mechanisms [1]–[3], or employing state space models (SSMs) [4], [5], these approaches enhance representational capacity and achieve impressive performance on several benchmark datasets. However, the pursuit of higher restoration accuracy often entails a dramatic surge in parameter count and computational cost. This heavy computational burden renders such complex models difficult to deploy on resource-constrained edge devices, such as drones and mobile terminals, in practical application scenarios.

Beyond computational inefficiency, existing restoration pipelines face inherent architectural bottlenecks. As illustrated in Figure 1, they generally fall into two sequential

Runci Bai is with the China Academy of Information and Communications Technology, Beijing 100080, China (email: 05201102@cumt.edu.cn).
Kui Jiang is with the Harbin Institute of Technology, Harbin 150000, China (email: jiangkui@hit.edu.cn).
Xiang Chen is with the Nanjing University of Science and Technology, Nanjing 210094, China (email: chenxiang@njust.edu.cn).
Chen Wu is with the National University of Defense Technology, Changsha 410073, China (email: wuchen5X@mail.ustc.edu.cn).
Dianjie Lu and Guijuan Zhang are with Shandong Normal University, Jinan 250358, China (email: {ludianjie, zhangguijuan}@sdnu.edu.cn).
Zhuoran Zheng is with Qilu University of Technology, Jinan 250353, China (email: zhengzr@njust.edu.cn).
Corresponding author: Zhuoran Zheng.

paradigms: 1) Detail enhancement followed by color rendition (Figure 1(a)). The enhancement process may alter the original color distribution, forcing the subsequent color module to adapt to modified pixel values. Over-sharpening in this stage often leads to unnatural color transitions and local distortions. 2) Color rendition followed by detail enhancement (Figure 1(b)). The smoothing nature of color mapping attenuates original edge and texture details, complicating subsequent high-frequency restoration. Consequently, detail enhancement algorithms may excessively boost local contrast, disrupting the finely adjusted color balance and causing oversaturation or color casts.

To circumvent these sequential limitations, LUTwithBGrid [6] proposed fusing 3D LUTs with bilateral grids to incorporate spatial information in a unified manner. While promising, this method neglects the correlation between color and edge features, limiting the synergy between color mapping and edge preservation. More critically, LUTwithBGrid relies on standard trilinear interpolation, applying a fixed, isotropic kernel to all pixels regardless of local image geometry. This uniform sampling across edges, textures, and flat regions overlooks the curved manifold of natural images, inevitably introducing edge blur and ghosting artifacts.

To address these issues, we propose $6^{\text{th}}$Grid-Net, a unified and lightweight remote sensing image restoration framework. $6^{\text{th}}$Grid-Net seamlessly integrates the color rendition capability of 3D LUTs and the edge-preserving properties of bilateral grids into a novel six-dimensional fusion tensor. This architecture abandons the traditional sequential pipeline in favor of a parallel integrated design, coupling global color mapping with local detail enhancement in a single forward inference. Specifically, the first stage constructs a 6D tensor comprising color ($R$, $G$, $B$) and spatial-luminance ($X$, $Y$, $L$) dimensions, where each node stores enhancement values corresponding to specific spatial positions and color components. The second stage introduces a manifold-adaptive high-dimensional sampling mechanism to replace conventional trilinear interpolation. Unlike fixed-kernel interpolation, this mechanism dynamically adjusts the sampling kernel per pixel based on local edge orientation, texture strength, and color similarity. By elongating the kernel along edges to preserve continuity and suppressing it across edges to prevent blurring, it respects the intrinsic low-dimensional structure of natural images. Consequently, simultaneous color stylization and edge sharpening are achieved without multi-stage feature loss. Furthermore, we incorporate an edge-aware grid smoothing constraint and dynamic quantization to ensure robust artifact suppression and model compactness.

The proposed $6^{\text{th}}$Grid-Net is comprehensively evaluated on unified remote sensing datasets (including RICE [7] and SATEHAZE1K [8]), and its generalization capability is further validated on the ViCoW dataset [9]. It exhibits significant advantages over representative sequential methods and supports real-time processing of 1080P images on consumer-grade mobile devices. Our main **contributions** are summarized as follows:

- We propose $6^{\text{th}}$Grid-Net, a unified network that seamlessly integrates 3D LUT-based color rendition with bilateral grid-based detail preservation into a compact six-dimensional fusion tensor. This parallel architecture enables simultaneous global color stylization and local edge refinement in a single forward pass, eliminating the computational redundancy and error accumulation inherent in sequential pipelines.
- We introduce a manifold-adaptive high-dimensional sampling mechanism that dynamically adjusts the interpolation kernel based on local edge orientation, texture strength, and color similarity. Coupled with a novel edge-aware grid smoothing constraint, it effectively suppresses noise and ghosting artifacts while preserving edge sharpness, leading to highly natural enhancement outcomes.
- To facilitate practical deployment, we employ dynamic quantization to compress the model without compromising regression accuracy, resulting in a lightweight architecture tailored for resource-constrained edge devices. Extensive experiments demonstrate that our method achieves state-of-the-art restoration quality across various adverse weather conditions and exhibits strong generalization in recovering RGB colors from grayscale natural scene images.

## II. RELATED WORK

### A. Remote sensing image restoration

Recent advancements in remote sensing image restoration can be broadly categorized into two paradigms: discriminative methods (*e.g.*, state-space models and adversarial networks) and generative methods (*e.g.*, diffusion models).

Discriminative methods typically formulate restoration as a direct mapping from degraded to clean images, leveraging structured priors or adversarial training to achieve perceptually realistic results. Within this category, state-space models (SSMs) have recently shown great promise. For instance, MambaHR [10] addresses stray light interference in hyperspectral images by integrating a state-space module with channel attention, effectively capturing global and local spectral-spatial dependencies. Similarly, DACDM-CR [11] introduces a Dynamic Cloud-Adaptive Mamba module for SAR-assisted optical data cloud removal, reconstructing global contexts with linear complexity. Adversarial training is another prominent approach. Patch-GAN [12] integrates a masked autoencoder with a patch-based discriminator to reconstruct cloud-occluded regions via transfer learning. PM-DBGAN [13] employs a physical scattering model to guide a dual-branch GAN, enabling the unsupervised inversion of atmospheric degradation for thin cloud removal. Furthermore, MAGC-GAN [14] combines multi-scale adaptive graph convolutions with a GAN architecture to capture global contextual dependencies, recovering fine details in thick cloud-covered areas.

Conversely, generative and multi-temporal methods primarily rely on iterative denoising processes or temporal feature aggregation. DBD-CR [15] adopts a dual-branch diffusion residual framework, utilizing residual learning during the diffusion process to enhance cloud removal quality. GenDeg [16] synthesizes datasets containing various degradations via a latent diffusion model, training a unified restoration network with strong generalization. To address thick clouds, IRNet [17] employs an implicit reconstruction approach using multi-temporal information without relying on explicit cloud masks. SGDM [18] leverages a pre-trained generative prior and integrates structural semantics from vector maps to achieve large-factor super-resolution, mitigating semantic inaccuracy and texture blurring. GMDiT [19] extends the masked DiT architecture by capturing spatial and semantic relationships through graph prior modeling, thereby enhancing structural fidelity.

Despite these advances, inherent limitations persist. GAN-based approaches frequently suffer from training instability and mode collapse, compromising performance consistency across diverse cloud and haze conditions. Diffusion-based and multi-temporal methods, while achieving high fidelity, demand massive computational resources and prolonged inference times, precluding real-time or on-orbit deployment. Additionally, multi-temporal frameworks heavily depend on cloud-free auxiliary images—often unavailable in persistently cloudy regions. Most critically, many existing methods treat cloud removal as a pure mathematical reconstruction task, neglecting the physical atmospheric scattering processes and intrinsic spectral characteristics, which inevitably leads to spectral distortion in downstream applications.

### B. Unified frameworks for multi-degradation restoration

To address the "all-in-one" image restoration task, existing unified frameworks generally fall into three strategies: degradation-aware prompting, high-capacity network architectures, and generative modeling.

The first strategy explicitly models degradation types using contrastive learning or prompt mechanisms. AirNet [20] introduced contrastive learning with a degradation encoder to map different weather corruptions into separate feature spaces. Perceive-IR [21] employs a fine-grained quality perceiver for multi-level prompt learning, enabling superior perception of degradation severity. Yang et al. [22] utilize natural language descriptions as conditional prompts for unified processing, while Wu et al. [23] propose a lightweight prompt-based framework that dynamically guides reconstruction through degradation- and instance-level prompts, mitigating catastrophic forgetting. The second strategy employs powerful backbones to uniformly handle multiple degradations via multi-scale feature fusion or kernel modulation. TransWeather [24] adopts a Transformer-based single encoder-decoder architecture with window-based self-attention to simultaneously remove rain, fog, and snow. Cui et al. [25] propose a multi-scale kernel modulation module (comprising global, large-kernel, and local branches) to efficiently learn universal representations. The third strategy exploits generative models, offering strong theoretical unification by treating various degradations as the reverse process of a common generation procedure, as demonstrated by WeatherDiffusion [26] through iterative denoising.

Recently, emerging studies have begun to prioritize real-time efficiency for deployment on resource-constrained edge devices. RTEE-Net [27] integrates global brightness and local details via a holographic attention mechanism, achieving low-light enhancement in just 0.002 seconds with merely 50K parameters. LLF-LUT++ [28] combines Laplacian pyramid decomposition with image-adaptive 3D LUTs to deliver 4K image enhancement in 13 milliseconds.

Although these unified frameworks demonstrate strong multi-degradation capabilities, they present several drawbacks. Methods relying on predefined degradation priors often struggle to generalize to unseen complex weather conditions. Meanwhile, frameworks utilizing heavy backbones or iterative generative models incur prohibitive computational costs, rendering them unsuitable for resource-constrained platforms. While some recent methods achieve real-time processing, they often compromise restoration quality when faced with severe degradations. Most importantly, existing unified frameworks typically entangle color restoration with detail enhancement without explicitly modeling their interaction, leading to sub-optimal trade-offs between color fidelity and structural sharpness—a critical bottleneck that our proposed method aims to resolve.

## III. METHODOLOGY

### A. Overview

We propose $6^{th}$Grid-Net, as illustrated in Figure 2, an end-to-end method for remote sensing image dehazing that integrates global color mapping and local detail enhancement into a six-dimensional tensor, enabling collaborative optimization via a single high-dimensional interpolation. Given a hazy input image $\mathbf{I}$, it is first converted to linear RGB space, and its luminance map $\mathbf{LM}$ is computed. In the first stage, a lightweight U-Net predicts a weight grid $\mathbf{B}$, which is then used to construct a fusion tensor $\mathcal{T}$ that blends a pre-defined 3D LUT with original pixel information. In the second stage, each pixel is processed through manifold-adaptive high-dimensional sampling over this tensor to obtain the enhanced result.

### B. Bilateral Grid Weights Generation

Given the hazy input image $\mathbf{I}_{\text{lin}} \in \mathbb{R}^{3 \times H \times W}$ and its luminance map $\mathbf{L} \in \mathbb{R}^{H \times W}$, we aim to predict a bilateral grid weight $\mathbf{W} \in [0, 1]^{N_x \times N_y \times N_l}$ that controls the contribution of local details during fusion. Here, $N_x$ and $N_y$ are the spatial grid resolutions (set to $128 \times 72$ for 4K images), and $N_l$ is the number of luminance bins ($N_l = 24$).

As illustrated in the left branch of Figure 2, we employ a lightweight U-Net to generate $\mathbf{W}$. The U-Net consists of an encoder that progressively downsamples the input to capture multi-scale features, and a decoder that upsamples the features back to the target grid size. Skip connections concatenate encoder features with decoder features at corresponding scales, preserving spatial details. The final output layer uses a $3 \times 3$ convolution followed by a sigmoid activation to ensure the weights are bounded in $[0, 1]$.

Formally, the U-Net maps the input image to the grid weight:

$$\mathbf{W} = \text{UNet}(\mathbf{I}_{\text{lin}}), \tag{1}$$

where each entry $u_{i,j,k} \in [0, 1]$ represents the degree of local detail preservation at spatial location $(i, j)$ and luminance bin

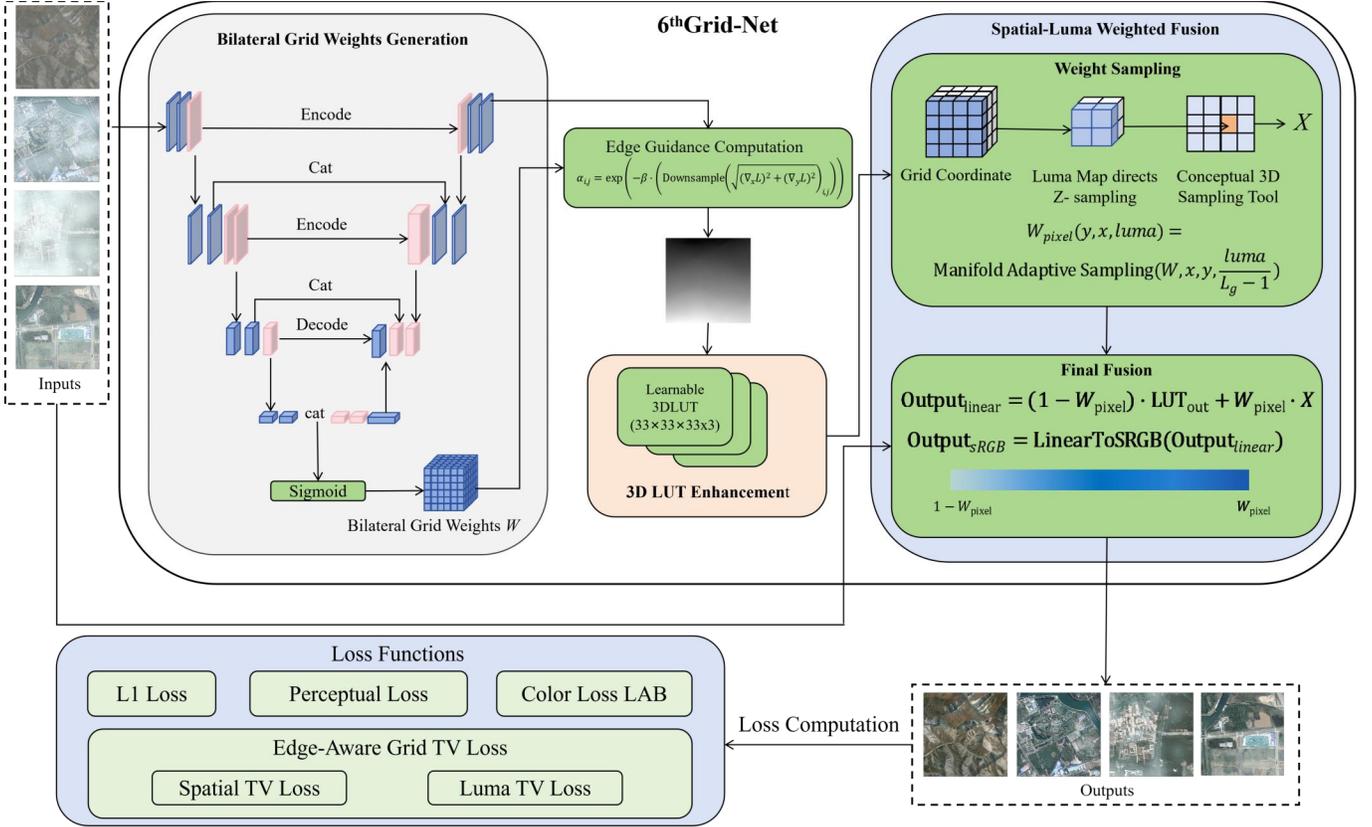

Fig. 2. Overall architecture of the proposed 6thGrid-Net, featuring a bilateral weight generation module and a spatial-luma weighted fusion mechanism. The bilateral weight grid is produced by an encoder–decoder network with skip connections, guided by an edge-aware term derived from the luminance map. Pixel-wise weights are then sampled via manifold-adaptive high-dimensional sampling based on spatial coordinates, luminance values, and local edge orientation, enabling adaptive fusion of the LUT-enhanced result with the original image. The model is optimized with a composite loss comprising L1, perceptual, color, and edge-aware TV terms.

$k$. A value of $w_{i,j,k} \approx 1$ indicates that the original pixel details are preserved, while $w_{i,j,k} \approx 0$ favors the global color mapping from the 3D LUT. This adaptive mechanism allows the model to selectively retain fine structures in textured regions while ensuring color consistency in smooth areas.

### C. Edge-Aware Smoothing Regularization

Directly learning the bilateral grid weight **W** from data may lead to undesirable local noise or blocky artifacts, as the network might exploit high-frequency variations to overfit training samples. To address this issue, we introduce an edge-aware total variation (TV) regularization that encourages **W** to vary smoothly in homogeneous regions while allowing sharp transitions near image edges.

We first compute the gradient magnitude of the input luminance map **L** to indicate edge strength. Using Sobel operators $\nabla_x$ and $\nabla_y$, the gradient magnitude $G \in \mathbb{R}^{H \times W}$ is obtained as:

$$G = \sqrt{(\nabla_x \mathbf{L})^2 + (\nabla_y \mathbf{L})^2}. \tag{2}$$

Since the grid weight **W** has a lower spatial resolution $N_x \times N_y$, we downsample $G$ to match the grid dimensions via bilinear interpolation, yielding $G_{\text{grid}} \in \mathbb{R}^{N_x \times N_y}$. The result is then replicated across the $N_l$ luminance bins to form $G_{\text{grid}} \in \mathbb{R}^{N_x \times N_y \times N_l}$, as edges are independent of the luminance layer.

To attenuate regularization at edges, we define an edge-aware weighting factor:

$$\alpha_{i,j,k} = \exp(-\beta \cdot G_{\text{grid}}(i,j,k)), \tag{3}$$

where $\beta > 0$ controls the decay rate. In flat regions ($G_{\text{grid}} \approx 0$), $\alpha \approx 1$ imposes strong smoothness; near strong edges ($G_{\text{grid}}$ large), $\alpha \approx 0$ relaxes the penalty, allowing weight changes to preserve edge sharpness.

The spatial TV term penalizes differences between neighboring grid nodes in the spatial dimensions:

$$R_{\text{spatial}} = \sum_{i,j,k} \alpha_{i,j,k} \left( |W_{i+1,j,k} - W_{i,j,k}| + |W_{i,j+1,k} - W_{i,j,k}| \right), \tag{4}$$

where the sums run over valid indices. This term is weighted by $\alpha$ to encourage smoothness in flat areas while preserving edges.

Additionally, we enforce smoothness along the luminance dimension to avoid abrupt transitions between adjacent luminance layers:

$$R_{\text{luma}} = \sum_{i,j,k} |W_{i,j,k+1} - W_{i,j,k}|, \tag{5}$$

with $k < N_l - 1$. No edge guidance is needed here, as luminance variations are inherently smooth.

The overall regularization loss combines both terms:

$$L_{\text{TV}} = \lambda_s R_{\text{spatial}} + \lambda_l R_{\text{luma}}, \quad (6)$$

where $\lambda_s$ and $\lambda_l$ are scalar weights (set to $0.01$ in practice). By incorporating $L_{\text{TV}}$ into the training objective, the network learns to produce weight grids that are spatially coherent in smooth regions yet adaptive to image structures, effectively suppressing artifacts while maintaining sharp details.

### D. Construction of Six-Dimensional Fusion Tensor

Having obtained the bilateral weight grid $W$ and the learnable 3D LUT $\text{LUT} \in \mathbb{R}^{N_c \times N_c \times N_c \times 3}$, we now construct a six-dimensional fusion tensor $T$ that jointly encodes global color mapping and spatially varying detail preservation. The tensor $T$ is defined over six dimensions: spatial coordinates $X$ and $Y$, luminance level $L$, and RGB color indices $R$, $G$, $B$. Its shape is $\mathbb{R}^{N_x \times N_y \times N_l \times N_c \times N_c \times N_c \times 3}$, where $N_x$, $N_y$ denote the spatial grid resolutions, $N_l$ the number of luminance bins, and $N_c$ the 3D LUT size. The final dimension stores the output linear RGB value for each combination of indices.

To balance computational efficiency and restoration quality, we set $N_x = 128$, $N_y = 72$, $N_l = 24$, and $N_c = 33$ for high-resolution remote sensing images. The spatial downsampling factor (approximately $30\times$) reduces the tensor size to about 1.935 million parameters, while the luminance dimension and LUT size provide sufficient granularity for accurate color styling and detail adaptation.

The tensor is populated by blending two information sources at each grid node: the global color mapping from the 3D LUT and the original pixel values from the input image. For a given grid node with spatial indices $(i, j)$ and luminance index $k$, we first sample the linear RGB image $I_{\text{lin}}$ at the center of the corresponding grid cell, obtaining $I_{\text{sample}}(i, j) \in \mathbb{R}^3$. The bilateral weight $w_{i,j,k} \in [0, 1]$ from $W$ controls the contribution of local details. The tensor entry for color indices $(r, g, b)$ is then defined as:

$$T(i, j, k, r, g, b) = (1 - w_{i,j,k}) \cdot \text{LUT}(r, g, b) + w_{i,j,k} \cdot I_{\text{sample}}(i, j). \quad (7)$$

This formulation ensures that in regions where $w_{i,j,k}$ is high (e.g., textured areas), the tensor retains the original pixel details; where $w_{i,j,k}$ is low (e.g., smooth areas), the tensor relies more on the global LUT mapping, thereby preserving color consistency. The entire construction process is fully differentiable, enabling end-to-end training. The resulting six-dimensional tensor $T$ compactly encapsulates both global and local information, forming the unified representation for the subsequent manifold-adaptive sampling step.

### E. Manifold-Adaptive High-Dimensional Sampling

Given the constructed six-dimensional fusion tensor $T$, the final enhanced output for each pixel is obtained through a manifold-adaptive high-dimensional sampling mechanism that dynamically adjusts the interpolation kernel based on local edge orientation, texture strength, and color similarity. For a pixel with spatial coordinates $(x, y)$, linear RGB values $(r, g, b)$, and luminance $l$ (computed as $l = 0.299r + 0.587g + 0.114b$), we first compute its continuous indices in each dimension:

$$x_g = \frac{x}{W} \cdot (N_x - 1), \quad (8a)$$
$$y_g = \frac{y}{H} \cdot (N_y - 1), \quad (8b)$$
$$l_g = l \cdot (N_l - 1), \quad (8c)$$
$$r_g = r \cdot (N_c - 1), \quad (8d)$$
$$g_g = g \cdot (N_c - 1), \quad (8e)$$
$$b_g = b \cdot (N_c - 1), \quad (8f)$$

where $W$ and $H$ are the image width and height, respectively. These real-valued indices are then decomposed into integer and fractional parts as in standard interpolation.

Unlike conventional trilinear interpolation, which always uses the eight surrounding vertices in the spatial-luminance dimensions, our manifold-adaptive scheme introduces three key improvements: (1) local tangent space estimation, (2) adaptive neighbor selection based on edge orientation, and (3) anisotropic weighting that respects the image manifold.

*1) Local Tangent Space Estimation:* From the luminance map $L$, we compute the gradient angle $\theta = \arctan(\partial L/\partial y, \partial L/\partial x)$ and edge strength $e = \|\nabla L\|$. The edge-aware uncertainty $u = \exp(-\beta e)$ from Eq. 3 is reused. The anisotropic scaling factor along the edge direction is defined as $\gamma = 1 + \kappa \cdot u$, where $\kappa$ controls the elongation (set to $3.0$ in practice). This factor allows the sampling kernel to stretch along the edge while remaining compact across it.

*2) Adaptive Neighbor Selection:* Instead of always using all eight spatial-luminance corners $(x_{0/1}, y_{0/1}, l_{0/1})$, we compute a confidence mask for each corner based on whether the line segment from the pixel to that corner crosses a strong edge. For a corner with spatial offset $(\Delta x, \Delta y)$, we evaluate the maximum gradient magnitude along the line segment. If the maximum exceeds a threshold $\tau$ (set to $0.15$), the corner is excluded. Typically, for a pixel on an edge, only 4–6 corners are retained, all lying on the same side of the edge. This prevents interpolation from mixing pixels from different objects, eliminating ghosting artifacts.

*3) Manifold Weight Computation:* For each selected corner $c$ with spatial-luminance coordinates $(x_c, y_c, l_c)$ and corresponding tensor value $T_c$, we compute the sampling weight as:

$$w_c = \exp\left(-\frac{\|\Delta \mathbf{p}_c\|^2}{2\sigma_s^2}\right) \cdot \exp\left(-\frac{\|\Delta \mathbf{rgb}_c\|^2}{2\sigma_r^2}\right) \cdot (1 - u_c), \quad (9)$$

where $\Delta \mathbf{p}_c = ((x_c - x_g)/\gamma_x, (y_c - y_g)/\gamma_y, l_c - l_g)$ is the spatial-luminance distance anisotropically scaled by $\gamma$ along the edge direction, $\Delta \mathbf{rgb}_c = (r_c - r_g, g_c - g_g, b_c - b_g)$ is the RGB difference, $\sigma_s = 0.5$ and $\sigma_r = 0.2$ are bandwidth parameters, and $u_c$ is the edge uncertainty at the corner. The weights are normalized: $\tilde{w}_c = w_c / \sum_c w_c$.

*4) Color Subspace Interpolation and Final Aggregation:* For each selected corner $c$, we first perform trilinear interpolation over the RGB dimensions using the eight surrounding

JOURNAL OF LATEX CLASS FILES, VOL. 14, NO. 8, AUGUST 2021 6LUT vertices (same as Eq. 8) to obtain an intermediate color $c_c$. Then the final pixel value is the weighted sum:

$$c_{xy} = \sum_c \tilde{w}_c \cdot c_c. \quad (10)$$

The resulting value $c_{xy}$ is the enhanced linear RGB value for the pixel at $(x, y)$. It is then converted back to the sRGB color space using the inverse of the linearization function, yielding the final dehazed pixel. This manifold-adaptive sampling is fully differentiable and adds only a small computational overhead (approximately 0.05 ms per 1080p image) because the neighbor selection and weight computation are implemented via efficient grid-based pre-filtering.

*5) Training Objectives:* To train $6^{th}$Grid-Net, we adopt a composite loss function that jointly optimizes pixel-wise accuracy, perceptual quality, color fidelity, and grid smoothness. The total loss is defined as:

$$L_{total} = \lambda_{l1} L_{l1} + \lambda_{perc} L_{perc} + \lambda_{col} L_{col} + \lambda_{tv} L_{TV}, \quad (11)$$

where $\lambda_{l1}$, $\lambda_{perc}$, $\lambda_{col}$, and $\lambda_{tv}$ are balancing hyperparameters (set to 1.0, 0.1, 0.1, and 0.01, respectively).

L1 Loss $L_{l1}$: Computed in the linear RGB space between the predicted output $\hat{I}_{lin}$ and the ground truth $I_{lin}^{gt}$, encouraging precise pixel-level reconstruction:

$$L_{l1} = \frac{1}{N} \sum \left\| \hat{I}_{lin} - I_{lin}^{gt} \right\|_1, \quad (12)$$

Perceptual Loss $L_{perc}$: Applied in the sRGB space using a pre-trained VGG-16 network to measure feature-level similarity, improving visual quality:

$$L_{perc} = \frac{1}{C} \sum_l \left\| \phi_l(\hat{I}_{srgb}) - \phi_l(I_{srgb}^{gt}) \right\|_1. \quad (13)$$

where $\phi_l$ denotes the feature map from the $l$-th layer of VGG-16.

Color Loss $L_{col}$: Defined in the LAB color space to enhance color accuracy. The L1 distance is computed separately for the luminance ($L$) and chrominance ($a, b$) channels:

$$L_{col} = \| L_{out} - L_{gt} \|_1 + \| a_{out} - a_{gt} \|_1 + \| b_{out} - b_{gt} \|_1. \quad (14)$$

TV Regularization Loss $L_{TV}$: Enforces smoothness on the bilateral weight grid $W$ while preserving edges, as detailed in Section III-C. It combines spatial and luminance total variation terms:

$$L_{TV} = \lambda_s R_{spatial} + \lambda_l R_{luma}, \quad (15)$$

where $R_{spatial}$ incorporates edge-aware weighting to prevent over-smoothing at boundaries, and $R_{luma}$ ensures coherent variation across luminance bins.

The composite loss is minimized end-to-end using the Adam optimizer, enabling the network to simultaneously learn the 3D LUT, the weight generation UNet, and the fusion tensor parameters, thereby achieving high-quality dehazing with both global color consistency and sharp local details.

To more clearly illustrate the computational procedure of the proposed manifold-adaptive high-dimensional sampling, Algorithm ?? presents a PyTorch-style pseudo-code implementation of its core steps.

TABLE I
PSEUDO-CODE OF MANIFOLD-ADAPTIVE HIGH-DIMENSIONAL SAMPLING (PYTORCH STYLE)

| Step | Operation | Description |
|---|---|---|
| 1 | Input: hazy image, 3D LUT, bilateral weight grid, grid parameters | img, lut, W, (Nx,Ny,Nl,Nc) |
| 2 | Compute continuous indices (spatial, luminance, RGB) | xg = x/W * (Nx-1); yg = y/H* (Ny-1); lg = l* (Nl-1); rg = r* (Nc-1); ... |
| 3 | Estimate local tangent space (edge orientation, strength, uncertainty) and anisotropic factor | theta, e = grad(lum); u = exp(-beta*e); gamma = 1 + kappa*u |
| 4 | Adaptive neighbor selection: keep corners on same side of edge | mask = [] for corner in corners: if max_grad_along_segment(corner) < tau: mask.append(corner) |
| 5 | Compute manifold weight for each selected corner (spatial distance + RGB similarity + edge uncertainty) | w_c = exp(- \|\| dp \|\| **2/(2*sigma_s**2)) * exp(- \|\| drgb \|\| **2/(2*sigma_r**2)) * (1-u_c) |
| 6 | Trilinear interpolation in RGB subspace per corner | c_c = trilinear_interp(lut, rg, gg, bg) |
| 7 | Final aggregation: weighted sum | out = sum(w_c * c_c) / sum(w_c) |

## IV. EXPERIMENTAL

### A. Experimental Settings

*1) Datasets:* $6^{th}$Grid-Net is evaluated on a unified dataset comprising two public benchmarks, SateHaze1K and RICE. In addition, the color recovery capability of $6^{th}$Grid-Net has been validated on the public dataset ViCoW.

SateHaze1K is a publicly available dataset for remote sensing image dehazing. The dataset integrates multi-sensor data and consists of 1,200 pairs of 640×640 images, each pair containing a hazy RGB image and the corresponding clear ground truth. The images are categorized into three levels of haze concentration: thin, moderate, and thick, with 400 pairs in each category. For each category, the images are split into 320 pairs for training, 40 pairs for validation, and 40 pairs for testing.

The Remote sensing Image Cloud rEmoving(RICE) dataset is a publicly available dataset for cloud removal in remote sensing images. The dataset consists of two subsets: RICE-I and RICE-II. The RICE-I subset is selected, which contains 500 image pairs of 512×512 pixels, each comprising a cloud-contaminated image and its corresponding cloud-free ground truth. Of these, 400 pairs are used for training, 50 pairs for validation, and 50 pairs for testing.

**A unified dataset combining SateHaze1K and RICE was constructed**, consisting of 1,700 image pairs in total: 500 cloudy pairs from RICE, along with 400 thick haze, 400 moderate haze, and 400 thin haze pairs from SateHaze1K. Each pair includes a remote sensing image to be processed and its corresponding clear ground-truth image. Among these, 1,360 pairs were used for training, 170 pairs for validation, and 170 pairs for testing, with all images uniformly resized to $512 \times 512$.

The ViCoW [9] dataset is a collection of image pairs specifically designed for historical image restoration and colorization tasks, with a focus on 20th-century Vietnam War-era film footage. It contains a total of 1,896 high-resolution ($1280 \times 720$) image pairs, extracted from four historically significant Vietnamese films set during the Vietnam War era. The dataset is organized into training, validation, and test sets, enabling researchers to train and assess deep learning models for restoring and colorizing historical imagery. Each pair consists of an original color frame and its corresponding grayscale version, generated using the ITU-R BT.601 luminance formula.

*2) Implementation Detail:* $6^{th}$Grid-Net adopts an encoder-decoder architecture for bilateral weight grid generation. The U-Net comprises four encoding levels with channel dimensions 32, 64, 128, 256, each followed by a max-pooling layer. The decoder symmetrically upsamples the features and concatenates them with the corresponding encoder outputs via skip connections. The output of the final decoder layer is bilinearly interpolated to the target grid resolution of $N_x = 128$, $N_y = 72$, and $N_l = 24$, followed by a $3 \times 3$ convolution with Sigmoid activation to produce the grid weight $\mathbf{W} \in [0, 1]^{128 \times 72 \times 24}$. The 3D LUT is initialized as an identity mapping with size $N_c = 33$. The model is optimized using the Adam optimizer ($\beta_1 = 0.9$, $\beta_2 = 0.999$, weight decay = 0), with an initial learning rate of $1 \times 10^{-4}$. A cosine annealing scheduler is employed to gradually reduce the learning rate over 100 epochs, and the batch size is set to 16.

For the unified remote sensing dataset, the model is trained for 100 epochs. The total loss function is a weighted combination of four components:

$$L_{total} = \lambda_{l1} L_{l1} + \lambda_{perc} L_{perceptual} + \lambda_{color} L_{color} + \lambda_{TV} L_{TV}, \quad (16)$$

where $L_{l1}$ is the L1 loss computed in the linear RGB space, $L_{perceptual}$ is the perceptual loss based on VGG-16 features, $L_{color}$ is the color loss in the LAB space, and $L_{TV}$ is the edge-aware total variation regularization. The corresponding weights are set to $\lambda_{l1} = 1.0$, $\lambda_{perc} = 0.1$, $\lambda_{color} = 0.1$, and $\lambda_{TV} = 0.01$. The TV loss itself consists of spatial and luminance components, each with a coefficient of 1.0, and the edge decay parameter is set to $\beta = 10.0$.

Data preprocessing first converts the input sRGB images to the linear RGB space using the standard gamma-curve transformation, followed by luminance computation with the coefficients $(0.299, 0.587, 0.114)$. The only data augmentation applied during training is random cropping to $256 \times 256$ patches. During testing, images are kept at their original resolution. No learning rate warm-up or early stopping is used.

TABLE II
QUANTITATIVE COMPARISON OF DIFFERENT METHODS ON THE UNIFIED REMOTE SENSING DATASET.

| Method | PSNR↑ | SSIM↑ | LPIPS↓ | Params↓ | FLOPs↓ | Size↓ | FPS↑ |
|---|---|---|---|---|---|---|---|
| Diffusion [35] | 21.29 | 0.81 | 0.10 | 109.74 | 1979.12 | 418.61 | 0.44 |
| RSHazeDiff [36] | 17.67 | 0.83 | 0.17 | 0.63 | 53.02 | 2.41 | 13.31 |
| Restormer [37] | 15.25 | 0.24 | 0.78 | 119.12 | 340.80 | 1028.55 | 3.70 |
| AOD-Net [38] | 16.48 | 0.80 | 0.24 | 0.00 | 0.54 | 0.01 | 309.34 |
| HyperHazeOff [39] | 25.14 | 0.92 | 0.08 | 2.47 | 38.09 | 29.07 | 8.86 |
| LUTwithGrid [6] | 20.67 | 0.81 | 0.21 | 0.46 | 0.67 | 1.78 | 44.50 |
| CLUT-Net [40] | 20.83 | 0.80 | 0.20 | 0.29 | 0.08 | 1.12 | 166.26 |
| $6^{th}$Grid-Net (ours) | 22.52 | 0.84 | 0.16 | 1.94 | 8.09 | 7.84 | 70.83 |

All baseline methods are retrained under the same protocol to ensure a fair comparison. After training, the UNet weights are quantized to INT8 using dynamic post-training quantization, while the 3D LUT remains in FP32, resulting in a model size of under 2MB and enabling real-time inference on mobile devices.

*3) Evaluation Metrics:* To comprehensively evaluate the restoration performance of $6^{th}$Grid-Net, we adopt three full-reference metrics for quantitative analysis. Peak Signal-to-Noise Ratio (PSNR) [29] measures pixel-level reconstruction accuracy, with higher values indicating better fidelity to the ground truth. Structural Similarity Index (SSIM) [29] evaluates the preservation of brightness, contrast, and structural information, where higher values reflect superior structural consistency. Learned Perceptual Image Patch Similarity (LPIPS) [30] quantifies perceptual quality using deep feature embeddings; lower LPIPS values indicate results that are closer to human visual perception.

For efficiency evaluation, we report the following metrics. Number of parameters (Params) [31] reflects the model complexity, with fewer parameters implying a lighter architecture. FLOPs (floating point operations) [32] measures the computational cost during inference, where lower FLOPs enable faster processing. Model size [33] refers to the storage space occupied by the saved model file, which directly impacts deployment feasibility on resource-limited devices. Frames per second (FPS) [34] is measured on a mainstream mobile device to evaluate real-time capability, with higher FPS indicating better suitability for edge deployment.

### B. Experimental Results

To comprehensively evaluate the proposed 6thGrid-Net, we conduct two sets of experiments. The first set compares our method against state-of-the-art (SOTA) remote sensing image dehazing and all-in-one restoration methods on the unified dataset. The second set validates the generalization capability of 6thGrid-Net in a challenging color recovery task: restoring RGB colors from grayscale natural scene images using the ViCoW dataset. All baseline methods are retrained under the same protocol for fair comparison.

*1) Comparison with State-of-the-Art Methods:* We compare the proposed $6^{th}$Grid-Net against seven representative methods on the unified remote sensing dataset. Table II reports the quantitative results in terms of reconstruction fidelity, perceptual quality, and efficiency. Among all competitors, our method achieves the highest PSNR (22.52 dB) and SSIM (0.836), as well as the lowest LPIPS (0.164), indicating superior pixel-wise accuracy, structural preservation, and perceptual naturalness. Although some lightweight models (e.g., AOD-Net, CLUT-Net) offer faster inference or smaller model size, they lag behind in restoration quality, especially under thick haze or cloud conditions. In contrast, $6_{th}$Grid-Net strikes a favorable trade-off between quality and efficiency, with 1.935M parameters and 8.09G FLOPs, while still running at 70.8 FPS on a mobile device.

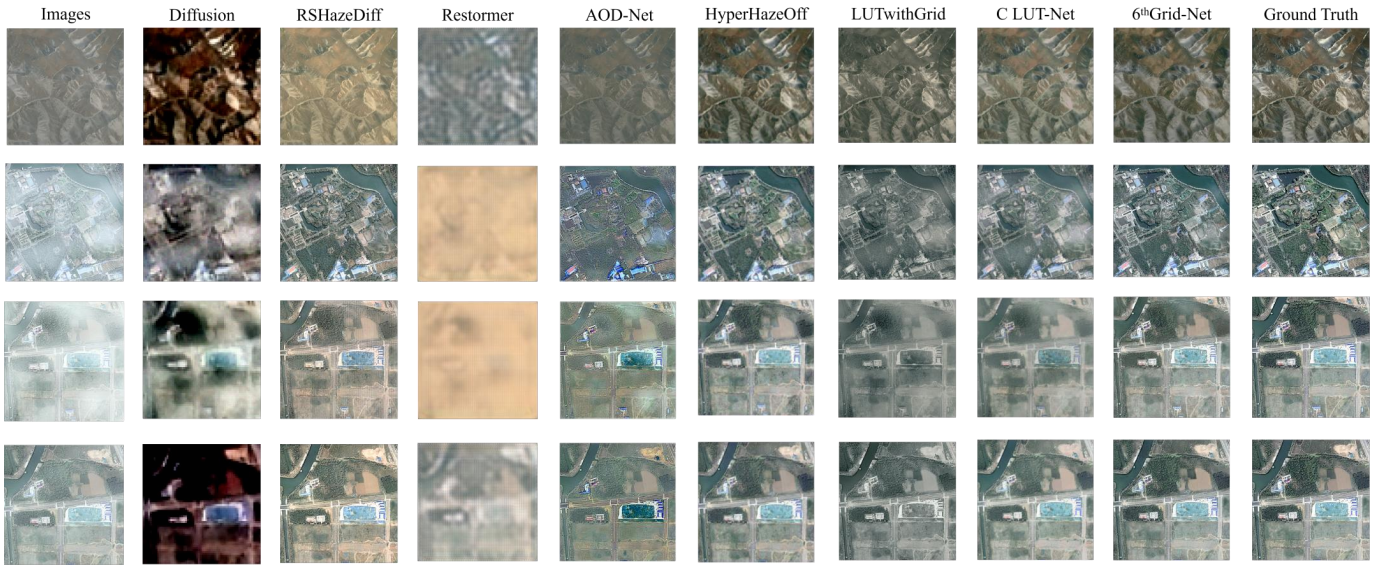

Fig. 3. **Qualitative results on remote sensing image dehazing.** Rows from top to bottom correspond to four degradation types: cloudy, moderate haze, thick haze, and thin haze, respectively. Columns from left to right show the input hazy image, the dehazed results produced by different baseline methods, the output of our proposed $6_{th}$Grid-Net (second last column), and the ground-truth clear image (last column). Compared with other methods, our $6_{th}$Grid-Net generates clearer textures, more faithful color restoration, fewer artifacts, and better edge preservation across all degradation levels.

TABLE III
COLOR RECOVERY PERFORMANCE ON THE VICOW DATASET.

| Method | PSNR ↑ | SSIM ↑ | LPIPS ↓ |
|---|---|---|---|
| AOD-Net | 28.10 | 0.96 | 0.14 |
| $6_{th}$Grid-Net | **28.61** | **0.97** | **0.13** |

Fig. 3 provides a qualitative comparison. It can be observed that methods such as Diffusion and RSHazeDiff produce oversmoothed outputs with residual haze, while Restormer and AOD-Net suffer from severe color distortion and blurring. LUTwithGrid and CLUT-Net recover reasonable colors but fail to preserve fine edges and textures. In comparison, $6_{th}$Grid-Net consistently reconstructs the sharpest details, most vivid colors, and cleanest backgrounds, effectively removing both thin and thick haze without introducing artifacts. These results validate the advantage of our manifold-adaptive high-dimensional sampling mechanism in jointly optimizing global color rendition and local edge preservation.

*2) Color Recovery from Grayscale Images:* To evaluate the generalization capability of our method beyond dehazing, we conduct color recovery experiments on the ViCoW dataset, which consists of grayscale-to-RGB image pairs extracted from historical Vietnam War-era film footage. Table III reports the quantitative results. Our method achieves a PSNR of 28.61 dB, an SSIM of 0.9653, and an LPIPS of 0.1326, indicating high-fidelity color reconstruction with strong structural and perceptual similarity to the ground truth.

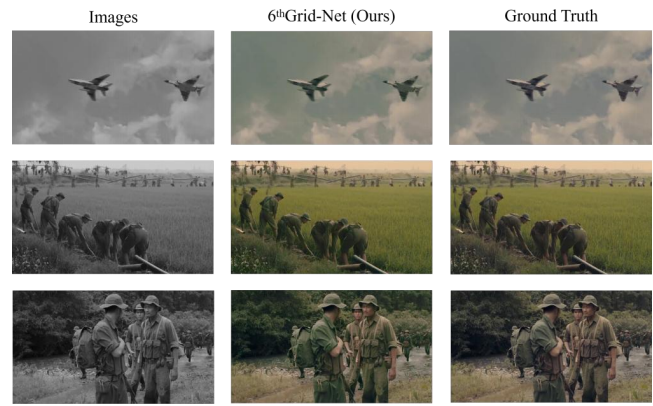

Fig. 4. **Color recovery results on the ViCoW dataset.** $6_{th}$Grid-Net reconstructs vivid and semantically plausible colors, closely matching the ground truth while preserving fine structural details.

Fig. 4 provides a qualitative comparison. For each of the three representative samples, the first column shows the input grayscale image, the second column presents the colorized output of our $6_{th}$Grid-Net, and the third column is the ground-truth color frame. It can be observed that our model not only recovers semantically plausible colors but also preserves fine textures and edges without introducing noticeable artifacts. The colorized results are visually close to the ground truth, demonstrating that the manifold-adaptive high-dimensional sampling mechanism, originally designed for dehazing, generalizes well to the task of grayscale image colorization. These results further confirm the versatility of our unified framework in handling diverse image restoration tasks.

TABLE IV
QUANTITATIVE COMPARISON BEFORE AND AFTER DYNAMIC QUANTIZATION.

| Model | PSNR↑ | SSIM↑ | LPIPS↓ | Params↓ | FLOPs↓ | Size↓ | FPS↑ |
|---|---|---|---|---|---|---|---|
| Pre-quant. (FP32) | 22.52 | 0.84 | 0.16 | 1.94 | 42.01 | 23.48 | 70.56 |
| Post-quant. (INT8) | 22.52 | 0.84 | 0.16 | 1.94 | **8.10** | **7.84** | **70.83** |

TABLE V
ABLATION STUDY OF KEY COMPONENTS. PARAMS, FLOPs, AND MODEL SIZE REMAIN CONSTANT (1.935M, 8.092G, 7.84MB) ACROSS ALL VARIANTS.

| Variant | M-S | E-TV | BW | PSNR↑ | SSIM↑ | LPIPS↓ | FPS↑ |
|---|---|---|---|---|---|---|---|
| a | ✗ | ✓ | ✓ | 21.98 | 0.82 | 0.19 | 61.61 |
| b | ✓ | ✗ | ✓ | 21.95 | 0.812 | 0.19 | 60.82 |
| c | ✗ | ✗ | ✓ | 20.67 | 0.82 | 0.17 | 59.13 |
| d | ✓ | ✓ | ✗ | 18.35 | 0.77 | 0.20 | 61.27 |
| Ours | ✓ | ✓ | ✓ | **22.52** | **0.84** | **0.16** | **70.83** |

*C. Ablation Studies*

*1) Quantitative ablation experiment:* To further validate the deployability of our method on resource-constrained edge devices, we apply dynamic post-training quantization to the UNet weights (INT8) while keeping the 3D LUT in FP32. Table IV compares the model before and after quantization. Notably, the quantized model retains exactly the same reconstruction quality (PSNR: 22.52 dB, SSIM: 0.8360, LPIPS: 0.1644) as its full-precision counterpart, while reducing the FLOPs from 42.007 G to 8.092 G (approximately 5.2× reduction) and the model size from 23.48 MB to 7.84 MB (approximately 3.0× reduction). Moreover, the inference speed slightly improves from 70.56 FPS to 70.83 FPS on a mobile device. These results demonstrate that our quantization strategy effectively compresses the model without sacrificing accuracy, making 6$^{\text{th}}$ Grid-Net well-suited for real-time remote sensing image dehazing on edge platforms.

*2) Component ablation experiment:* To validate each component, we compare five variants on the unified dataset (Table V). Variant (a) disables manifold-adaptive sampling (M-S); (b) removes edge-aware TV regularization (E-TV); (c) disables M-S, E-TV and replaces the bilateral weight grid (BW) with a fixed uniform weight; (d) removes BW; and Ours includes all three components. All variants share identical parameters, FLOPs and model size. As shown in Table V, removing M-S (variant a) reduces PSNR from 22.52 dB to 21.98 dB and increases LPIPS to 0.1857, confirming that M-S avoids cross-edge blur. Variant (b) (without E-TV) yields lower PSNR (21.95 dB) and higher LPIPS (0.1869), indicating that E-TV suppresses weight noise. Variant (c) performs significantly worse (PSNR 20.67 dB, LPIPS 0.1746) due to the absence of all adaptive mechanisms. Most critically, variant (d) (without BW) suffers the largest drop (PSNR 18.35 dB, SSIM 0.7739, LPIPS 0.2044), proving that the bilateral weight grid is indispensable for spatially varying fusion. The full model achieves the best performance across all metrics (PSNR 22.52 dB, SSIM 0.8360, LPIPS 0.1644) and the highest FPS (70.83). Thus, all three components contribute positively, with BW being the most critical.

## V. DISCUSSION

*A. Visualization and Analysis of Bilateral Weight Grid*

The core idea of the bilateral weight grid is to enable spatially-adaptive fusion between global LUT-based color mapping and original pixel details, thereby overcoming the mutual interference problem inherent in sequential pipelines (Fig. 6). Instead of applying a fixed blending factor across the entire image, our method learns a weight grid $\mathbf{W} \in [0, 1]^{N_x \times N_y \times N_l}$ that controls, for each spatial location and luminance level, how much of the original detail should be preserved. As shown in Fig. 6, the weight heatmap (middle) and its overlay on the input image (right) clearly demonstrate that high weights (red) concentrate along edges and textured regions (e.g., building boundaries and vegetation), while low weights (blue) dominate in flat, homogeneous areas such as sky and haze. This behaviour directly addresses the limitation of sequential methods: in a "detail-then-color" pipeline, edge enhancement would distort the original color distribution, forcing the subsequent color rendition module to adapt to already-modified pixel values; conversely, a "color-then-detail" pipeline would have its edge-preserving capability weakened by the smoothing effect of color mapping. By learning a location- and luminance-dependent weight, our unified six-dimensional tensor decouples these two objectives: flat regions rely almost entirely on the LUT for consistent color styling, whereas detailed regions retain their original high-frequency information without being over-smoothed or artificially recolored.

The adaptability of the weight grid across luminance levels is further validated in Fig. 7. For a pixel located on a strong edge (e.g., building-sky boundary), the weight remains consistently high across all 24 luminance bins, ensuring that edge sharpness is preserved regardless of local brightness. In contrast, a pixel in a flat region (e.g., clear sky) maintains low weights throughout, indicating that the LUT alone provides sufficient color fidelity. A textured region (e.g., ground vegetation) exhibits moderate weights with only slight variation, reflecting that some local detail is beneficial but aggressive preservation is unnecessary. This consistent behaviour across luminance bins confirms that the model does not rely on a single brightness level to make blending decisions; instead, it learns a smooth, content-aware mapping along the luminance dimension, which avoids abrupt transitions that could otherwise introduce banding or halos.

Finally, Fig. 8 examines the spatial smoothness of the learned weight grid. The gradient magnitude map (right) reveals that the weight grid varies very little in homogeneous areas—e.g., the sky region appears nearly black—indicating that the edge-aware total variation regularization successfully suppresses noisy or blocky artifacts that would otherwise arise from unconstrained learning. Strong gradients only appear where the image itself contains salient edges (e.g., building contours), allowing the weight grid to change sharply exactly where detail preservation is needed. Without this regularization (ablated in Table V), the weight grid would exhibit scattered high-frequency fluctuations, which could lead to unnatural textures or flickering in video applications. The proposed edge-guided smoothing constraint thus strikes a balance: it enforces spatial coherence in flat regions while preserving the ability to adapt to image structures, yielding a weight grid that is both interpretable and effective for high-quality dehazing.

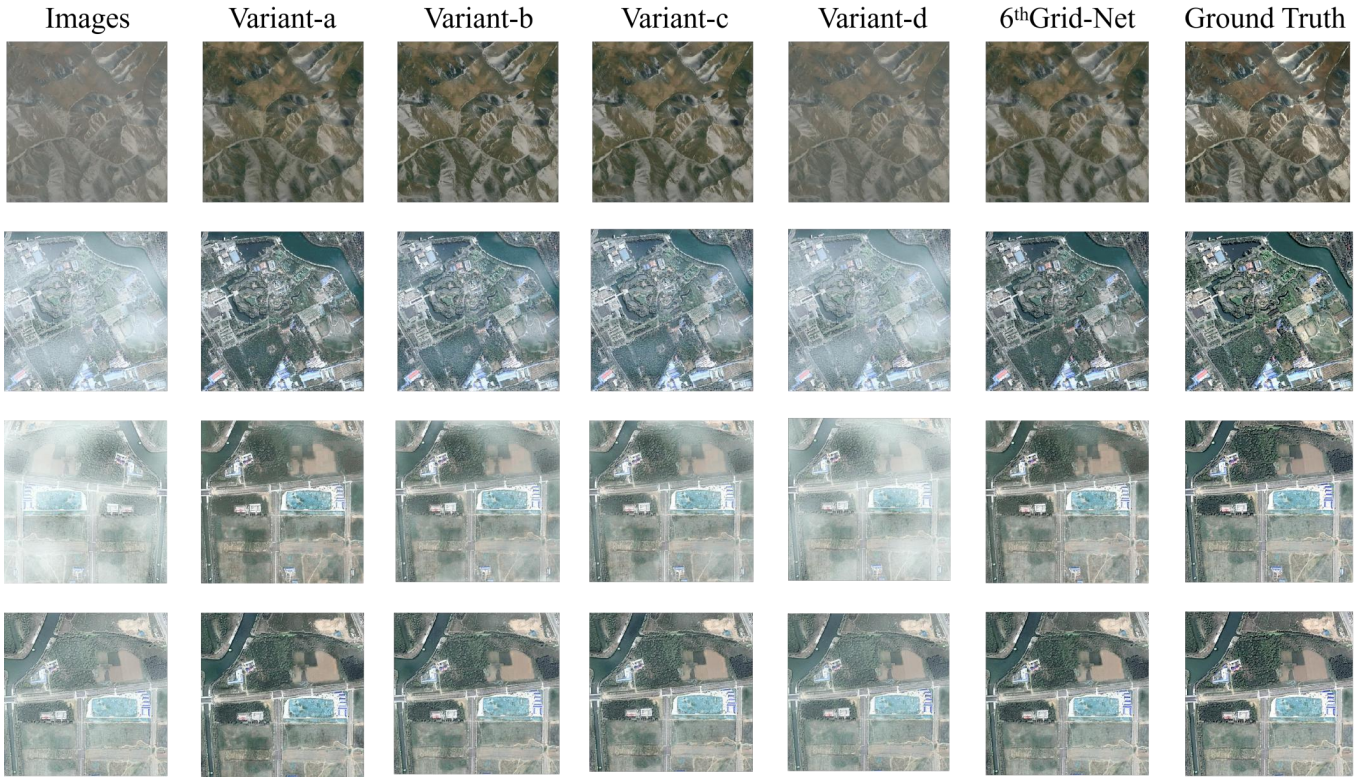

Fig. 5. **Ablation Results of Component Effectiveness.** Rows from top to bottom correspond to four degradation types: cloudy, moderate haze, thick haze, and thin haze, respectively. Columns from left to right show the input image, the processed results produced by different ablation variants (variants a–d as defined in Table V), the output of our complete model (second last column), and the ground-truth clear image (last column). Compared with the ablated variants, our full model achieves superior restoration with clearer textures, more faithful color restoration, fewer artifacts, and better edge preservation, confirming the necessity of each proposed component.

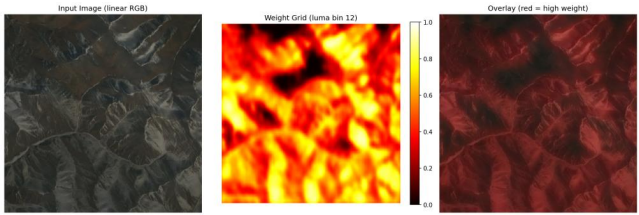

Fig. 6. **Visualization of the bilateral weight grid for a single luminance bin.** Left: input linear RGB image. Middle: weight heatmap (hot colormap) for the central luminance bin (index 12 out of 24), where red (value close to 1) indicates preservation of original details and blue (value close to 0) indicates reliance on global LUT mapping. Right: overlay of the weight heatmap on the input image, showing that high weights concentrate along edges and textured regions, while low weights dominate in flat areas.

### B. Analysis of Manifold-Adaptive Sampling

Conventional high-dimensional interpolation methods, such as trilinear interpolation, apply a fixed isotropic kernel to every pixel regardless of local image structure. While computationally efficient, this strategy inevitably mixes information from both sides of a strong edge, leading to blurred boundaries, halos, and color bleeding—artifacts that are particularly detrimental in remote sensing imagery where fine structural details (e.g., building outlines, road networks) must be preserved. To overcome this limitation, our method replaces the fixed kernel with a manifold-adaptive sampling mechanism that respects the intrinsic low-dimensional structure of natural images. Specifically, it estimates local edge orientation and strength from the luminance map, then dynamically adjusts the interpolation kernel: the kernel elongates along the edge direction to maintain continuity and suppresses weights across the edge to avoid mixing pixels from different objects. This design enables a single forward pass to simultaneously achieve global color stylization and local edge sharpening.

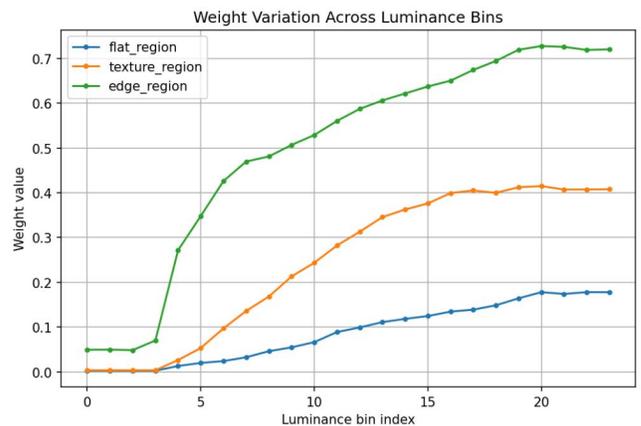

Fig. 7. **Weight values across luminance bins for three representative regions.** The curves correspond to a flat region, a textured region, and an edge region. The edge region maintains consistently high weights across all luminance levels to preserve structural details, whereas the flat region exhibits low weights throughout, indicating that the LUT-based color mapping is sufficient. The textured region shows moderate weights with a slight variation, demonstrating that the learned weight grid adapts to local image content.

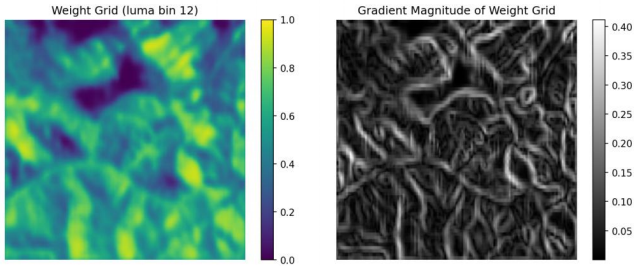

Fig. 8. **Smoothness analysis of the bilateral weight grid.** Left: weight map for the central luminance bin (viridis colormap). Right: gradient magnitude of the same weight map. The gradient is nearly zero in homogeneous areas, confirming that the edge-aware TV regularization effectively suppresses spurious noise and blocky artifacts. Strong gradients only appear along actual image edges, allowing the weight grid to change sharply where necessary to preserve fine structures.

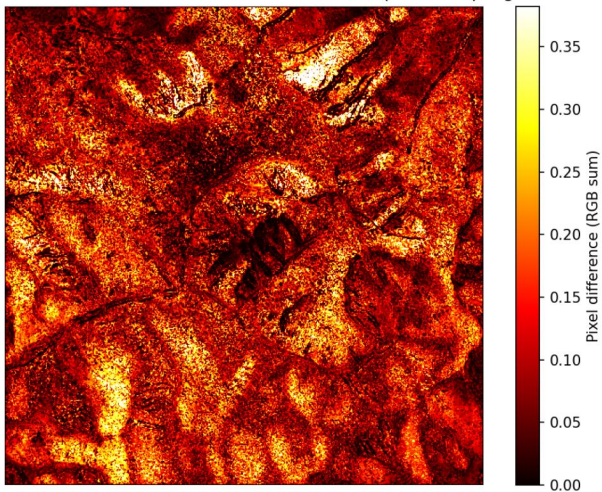

Fig. 9. **Difference map between standard and manifold-adaptive sampling.** The hot-colored map visualizes the per-pixel absolute difference (sum over RGB channels) between the outputs of standard trilinear interpolation and our manifold-adaptive sampling. Bright regions are concentrated along strong edges and textured areas, indicating that the adaptive mechanism primarily modifies the blending behavior where cross-edge mixing would otherwise occur, while leaving homogeneous regions nearly unchanged. This confirms that the manifold-adaptive sampling introduces targeted improvements without disturbing global color rendition.

Fig. 9 visualizes the spatial distribution of differences between the outputs of standard trilinear interpolation and our manifold-adaptive sampling. Bright regions in the hot-colored map correspond to locations where the two sampling strategies diverge most. Notably, these bright regions are predominantly concentrated along strong edges and fine-textured areas, whereas homogeneous regions (e.g., sky, calm water) exhibit nearly zero difference. This observation confirms that the adaptive mechanism introduces targeted modifications only where cross-edge mixing would otherwise occur, leaving the global color mapping in flat areas unchanged. Consequently, the model improves edge fidelity without compromising the color consistency achieved by the LUT-based component.

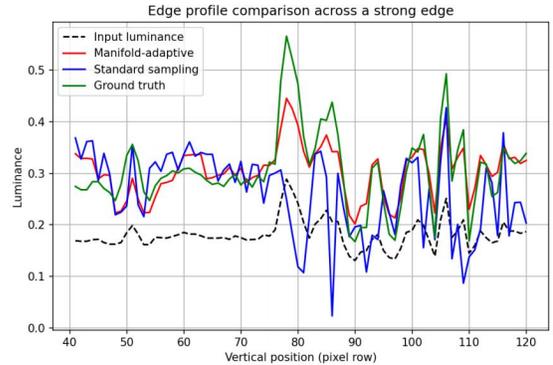

Fig. 10. **Edge profile comparison across a strong edge.** The plot shows luminance values along a vertical scanline crossing a salient edge. Compared with standard trilinear interpolation (blue), the proposed manifold-adaptive sampling (red) produces a substantially steeper transition that closely follows the ground truth (green), effectively suppressing cross-edge blur and preserving edge sharpness. The input luminance profile (black dashed) is also shown for reference.

Fig. 10 provides a quantitative comparison of luminance profiles along a vertical scanline that traverses a strong edge (e.g., a building-sky boundary). The input luminance (black dashed) shows an abrupt transition, characteristic of a sharp edge. Standard trilinear interpolation (blue) produces a much shallower slope, effectively blurring the edge and creating a gradual ramp that spreads across several pixels—a direct manifestation of cross-edge mixing. In contrast, our manifold-adaptive sampling (red) yields a substantially steeper transition that closely follows the input profile. The preservation of edge steepness directly translates into sharper boundaries and reduced halos in the final dehazed image.

In summary, the manifold-adaptive sampling mechanism addresses the fundamental shortcoming of fixed-kernel interpolation by explicitly respecting image edges. As evidenced by the difference map (Fig. 9) and the edge profile (Fig. 10), it effectively suppresses cross-edge blur while maintaining the ability to perform joint color rendition and detail enhancement in a single inference. The computational overhead remains negligible ($\approx 0.05$ ms per 1080p image), making the mechanism both effective and practical for real-time deployment on edge devices.

## VI. CONCLUSION AND FUTURE WORKS

In this paper, we propose $6^{th}$ Grid-Net, a unified remote sensing image dehazing framework that integrates LUT-based color rendition and bilateral grid-based edge preservation into a compact six-dimensional fusion tensor. By replacing conventional trilinear interpolation with a manifold-adaptive high-dimensional sampling mechanism that respects local edge orientation, texture strength, and color similarity, our method simultaneously achieves global color stylization and local detail enhancement in a single forward pass, avoiding the computational redundancy and feature loss of sequential pipelines. The introduced edge-aware grid smoothing regularization and dynamic post-training quantization further reduce model size by approximately $3\times$ without sacrificing reconstruction accuracy. Extensive experiments on the unified remote sensing dataset (SateHaze1K and RICE) demonstrate that $6^{\text{th}}$GridNet outperforms representative methods in terms of PSNR, SSIM, and LPIPS, while also exhibiting strong generalization to grayscale-to-RGB colorization on the ViCoW dataset. Moreover, the quantized model runs in real time on consumer-grade mobile devices, confirming its suitability for resource-constrained edge platforms. It should be noted, however, that the current six-dimensional tensor construction and manifold-adaptive sampling, though efficient, still involve a moderate memory footprint when scaling to ultra-high-resolution (e.g., 4K) imagery. Future work will explore adaptive grid pruning and learnable downsampling strategies to further reduce storage and computational costs while preserving edge fidelity and color accuracy.